\documentclass[11pt,a4paper]{article}

\usepackage[hyperref]{acl2020}
\usepackage{graphicx}
\usepackage{times}
\usepackage{latexsym}
\usepackage{mathabx}
\usepackage{natbib}
\usepackage{tablefootnote}
\usepackage[mathletters]{ucs}
\usepackage[utf8x]{inputenc}
\usepackage{footmisc}

\usepackage{microtype}

\aclfinalcopy 

\title{\textbf{Machine Translation Pre-training for Data-to-Text Generation - A Case Study in Czech}}

\author{\textbf{Mihir Kale} and \textbf{Scott Roy} \\
  Google \\
  \texttt{\{mihirkale,hsr\}@google.com}}

\date{}

\begin{document}
\maketitle
\begin{abstract}
While there is a large body of research studying deep learning methods for text generation from structured data, almost all of it focuses purely on English. In this paper, we study the effectiveness of machine translation based pre-training for data-to-text generation in non-English languages. Since the structured data is generally expressed in English, text generation into other languages involves elements of translation, transliteration and copying - elements already encoded in neural machine translation systems. Moreover, since data-to-text corpora are typically small, this task can benefit greatly from pre-training. Based on our experiments on Czech, a morphologically complex language, we find that pre-training lets us train end-to-end models with significantly improved performance, as judged by automatic metrics and human evaluation. We also show that this approach enjoys several desirable properties, including improved performance in low data scenarios and robustness to unseen slot values. 
\end{abstract}

\section{Introduction}
\begin{figure}
\noindent\includegraphics[width=0.5\textwidth]{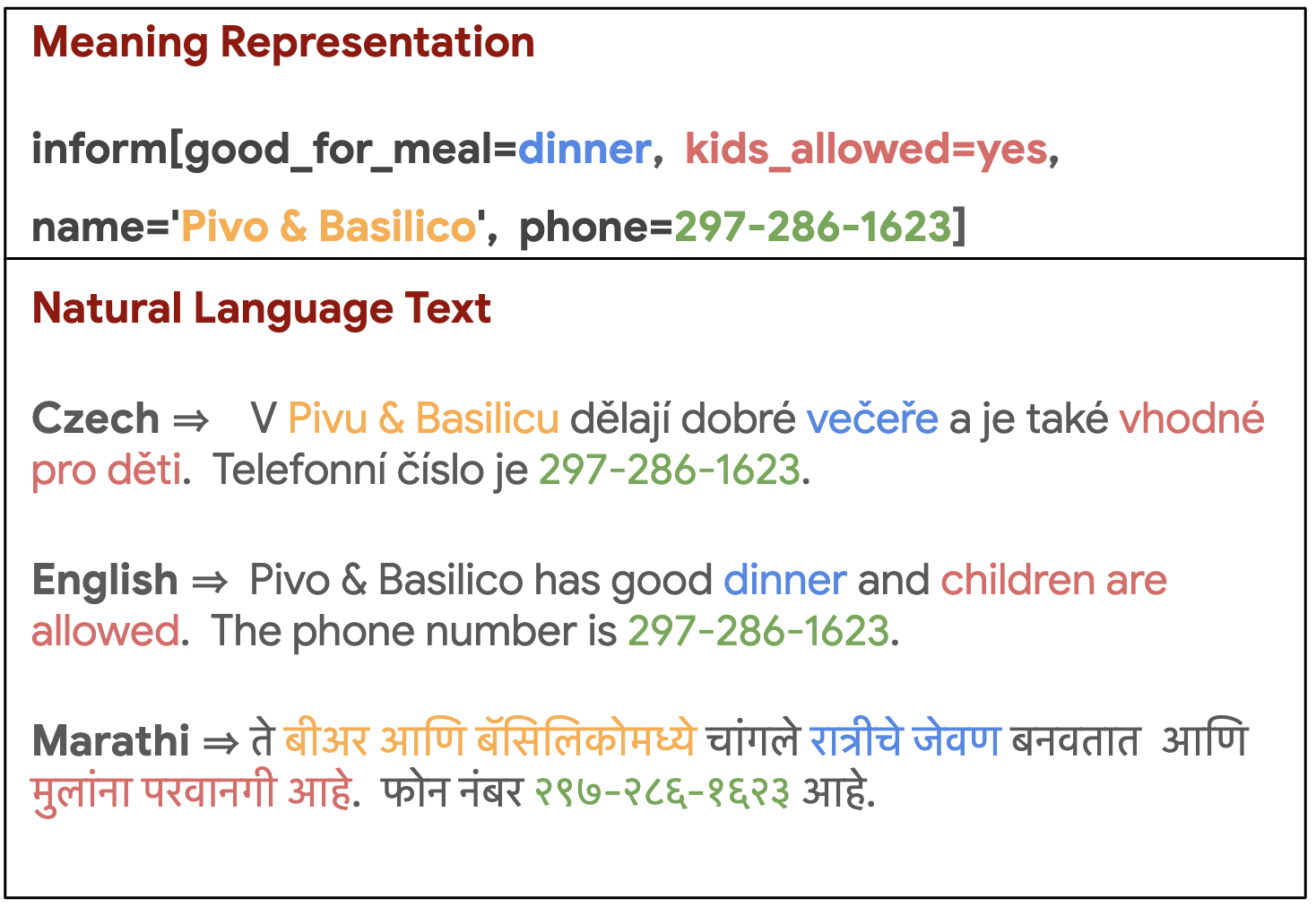}
\caption{
Generating text from structured data. Aligned segments from the structured data and natural language have the same color.
}
\label{fig:data-to-text}
\end{figure}

Data-to-Text refers to the process of generating accurate and fluent natural language text from structured data such as tables, lists, graphs etc.\citep{gatt2018survey}
It has several applications, including generating weather and sports summaries, response generation in task-oriented dialogue systems etc. For example, consider Figure \ref{fig:data-to-text}, in the context of a restaurant booking system. The system must take a meaning representation (MR) as input - in this case represented in the form of a dialogue act (\textsl{inform}) and a list of key value pairs related to the restaurant - and generate fluent text that is firmly grounded in the MR. \par

\begin{figure}
\noindent\includegraphics[width=0.5\textwidth]{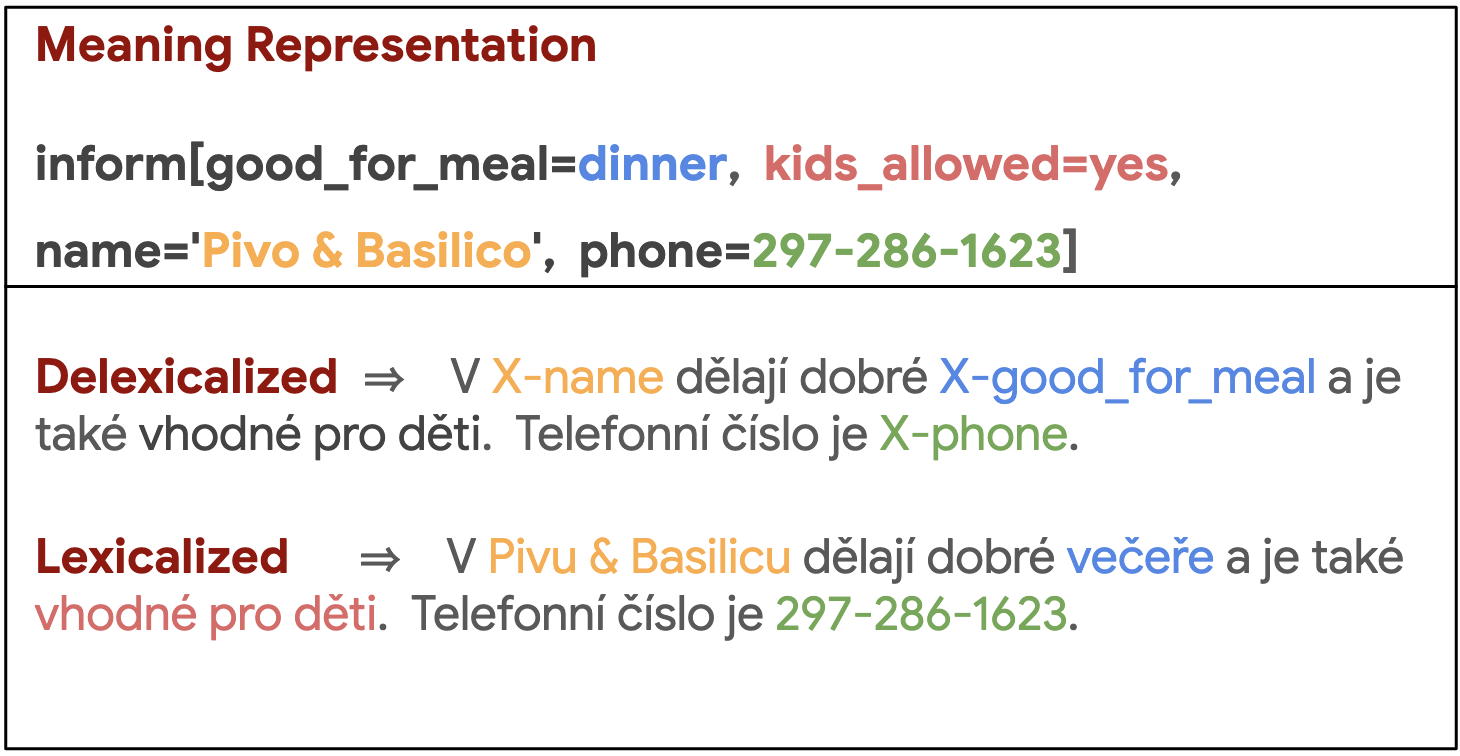}
\caption{
Example of lexicalized and delexicalized text, from the Czech NLG dataset.
}
\label{fig:delex}
\end{figure}

Data-to-text can broadly be classified into two categories with respect to the nature of the output text: lexicalized and delexicalized. Figure \ref{fig:delex}  provides an example of both. In the lexicalized setting, models are trained to produce the full natural text. We refer to these as lexicalized models. In the delexicalized setting, the slot-values are replaced with placeholders. Models are trained to produce output text with these placeholders. We refer to these as delexicalized models. The placeholders are filled in via a separate lexicalization step. For English, this is achieved by simply copying slot values from the structured data into the corresponding placeholders. \par
However, \citet{duvsek2019neural} recently highlighted the deficiencies of delexicalization and copy based methods in the presence of linguistic phenomena such as morphological inflection. For instance, in Figure \ref{fig:data-to-text}, when generating in Czech, the restaurant name "Pivo \& Basilico" from the MR must be correctly inflected to "Pivu \& Basilicu" to ensure fluency. Simple copying would fail \footnote{Nouns in Czech may have up to 14 different forms, depending on the context.}.  Moreover, in several languages, these complexities are compounded by the fact that inflecting a noun in a certain way requires changes to be made in the surrounding words (since modifying adjectives need to exhibit agreement).  This makes the lexicalization step complex, requiring extensive linguistic knowledge. Consequently, end-to-end systems that directly generate fully lexicalized text without depending on any external linguistic knowledge present an attractive alternative. However, their performance in terms of semantic accuracy tends to lag far behind their delexicalized counterparts, especially in the presence of slot values not seen during training. \citep{shimorina2018handling}. \par

In this work, we focus on generating text in non-English languages and show that it is possible to significantly reduce this accuracy gap by pre-training fully lexicalized models on an NMT task. For an example motivating the use of NMT, consider Figure \ref{fig:data-to-text} once again. In order to generate semantically correct and natural sounding text in Czech (Marathi), a data-to-text model would need to learn the following skills:
\begin{itemize}
    \item \textsl{Translate} the slot value "dinner" to the target language
    \item \textsl{Copy} the phone number correctly
    \item \textsl{Inflect} the restaurant name
\end{itemize}
In the case of Marathi, which has a different script, there is the additional challenge of \textsl{Transliterating} the restaurant name as well. \par
It is unreasonable to expect neural data-to-text models to learn all these skills, especially since the size of most NLG \footnote{While NLG is a broad term, in this paper, we use NLG and data-to-text interchangeably.} datasets is quite small. However, modern neural machine translation systems are already fairly adept at translating, transliterating, copying, inflecting etc. Consequently, we hypothesise that the parameters of a NMT model will act as a very strong prior for an NLG model. \par

\section{Related Work}
Earlier work on NLG was mainly studied rule-based pipelined methods, but recent works favor end-to-end neural approaches.  \citet{wen2015semantically} proposed the Semantically Controlled LSTM and were one of the first to show the success of neural networks for this problem, with applications to task oriented dialogue. Since then, some works have focused on alternative architectures - \citet{liu2018table} generate text by conditioning language models on tables, while \citet{puduppully2019data} propose to explictly model entities present in the structured data. The findings of the E2E challenge \citep{duvsek2018findings} show that standard seq2seq models with attention also perform well. \par
With the advent of ELMo, BERT \citep{devlin2018bert} and GPT-2 \citep{radford2019language}, the unsupervised pre-training + fine-tuning  paradigm has shown to be remarkably effective, leading to improvements in NLP tasks like classification, question answering and spoken language understanding \citep{siddhant2019unsupervised}. Results for generation tasks like summarization are also positive, albeit less dramatic. \citet{song2019mass} propose the MASS technique and obtain state-of-the-art results for summarization and unsupervised machine translation. \citet{freitag2018unsupervised} show that denoising autoencoders can be leveraged for unsupervised language generation from structured data. \citet{budzianowski2019hello} cast data-to-text as text-to-text generation and show that fine-tuning GPT language models can lead to performance competitive with architectures developed specifically for data-to-text. \citet{chen2019few} use language models to improve performance in the low resource scenario. \par
While the above works focus on unsupervised pre-training, \citet{siddhant2019evaluating} and \citet{schuster2018cross} examine transfer learning via neural machine translation for NLU tasks like spoken language understanding and named entity recognition in the cross-lingual setting. They find that the results are mixed and for several NLU tasks, unsupervised pre-training actually outperforms its NMT counterpart. Our work shows that for NLG, machine translation substantially outperforms unsupervised pre-training objectives. \par
Recently, \citet{chi2019cross} found multilingual pre-training techniques to be effective for cross-lingual language generation tasks like summarization and question generation. They focus on text-to-text NLG such as question generation and text summarization, where both the input and output are in the same language. In contrast, our work studies generation in the data-to-text setting, where the input is structured data as opposed to free form text and the output can be in any language. \par
The WNGT 2019 shared task provides a data-to-document dataset for German. However it is a small dataset that has been obtained by translating the English RotoWire dataset \citep{wiseman2017challenges}. Since the English dataset was automatically created by crawling and aligning sports score boxes and summaries,  large parts of the text in the RotoWire dataset are not grounded in the data. \citet{hayashi2019findings} find that techniques such as multilingual training, back-translation etc can help improve data-to-text performance in data scarce scenarios. Our focus is on NMT based transfer learning \footnote{We use pre-training and transfer learning interchangeably.} and it can be combined with all of the above techniques.

\section{Model Architecture}
We use the transformer \citep{vaswani2017attention} based encoder-decoder architecture by casting data-to-text as a seq2seq problem, where the structured data is flattened into a plain string consisting of a series of intents and slot key-value pairs. More exotic architectures have been suggested in prior work, but the findings of \citet{duvsek2018findings} show that simple seq2seq models are competitive alternatives, while being simpler to implement. Secondly, the transformer architecture is state-of-the art for NMT. Thirdly, keeping the pre-train and fine-tune architectures the same allows us to easily transfer knowledge between the two steps by parameter initialization.

\section{Pre-train + Fine-tune}
Our modeling approach is simple. We first use a parallel corpus to train a transformer based neural machine translation model that translates English text into the target language (Czech for our experiments). Next, we fine-tune this NMT model using a data-to-text corpus for a small number of steps. All the model parameters are updated in the fine-tuning process. \par

\section{Models and Baselines} \label{baselines}
\textbf{Machine Translation pre-training} This is our proposed approach (\textsl{nmt}), where we first train an NMT model and fine-tune it for the NLG task. We also experiment with fine-tuning a bidirectional machine translation (\textsl{binmt}), where the NMT model is trained to translate both from English to the target language and vice-versa. This translation is trained on the concatenation of English-Czech and Czech-English parallel data. \par
\textbf{Training from scratch} A baseline where all the parameters are learned from scratch, without any kind of transfer learning. This is a 1 layer Transformer model. Larger models trained from scratch did not improve performance. \par
\textbf{Unsupervised pre-training baseline} Monolingual data is generally far easier to obtain than bilingual data, which makes unsupervised pre-training techniques more attractive. Interestingly, \citet{wu2019beto} and \citet{pires2019multilingual} find that training multilingual BERT models on a combination of languages can lead to surprisingly effective cross-lingual performance on NLU tasks, without using any parallel data. Of the myriad unsupervised techniques, we choose MASS \citep{song2019mass} for our baseline since it has been shown to outperform other alternatives like BERT, left-to-right language models and denoising autoencoders for language generation tasks. We first train a unsupervised English-Czech MASS model and then fine-tune it for the NLG task. We denote this approach as  \textsl{mass}. From a transfer learning perspective, MASS is a state-of-the-art baseline. \par
\textbf{TGen} is a freely available open-source NLG system based on seq2seq + attention and was used as a strong baseline in the E2E challenge. \citet{duvsek2019neural} create a pipelined system consisting of : a TGen based model that outputs delexicalized text, a classifier that ranks the beam search hypotheses and a language model which which does the lexicalization by picking the exact surface form. We denote this combined system, consisting of all 3 components as \textsl{tgen-sota}. It is also currently the state-of-the-art for this dataset. Note that unlike \textsl{tgen-sota}, all our proposed models are trained to directly generate lexicalized outputs, which is a much harder task.
\section{Experimental Setup}

\begin{table}[]
\begin{tabular}{l|l}
\hline
Slot Type & Example Values         \\ 
\hline
name               & Kočár z Vídně, Green Spirit     \\ 
area               & Hradčany, Žatecká               \\ 
address            & Kaprova 38, Žatecká 30          \\ 
phone              & 250625609, 219289692            \\ 
good\_for\_meal    & lunch, dinner, breakfast        \\ 
near               & Powder Tower                    \\ 
food               & German, American                \\ 
price\_range       & cheap, expensive                \\ 
count              & 10, 21                          \\ 
price              & between 180 and 730 Kč \\ 
postcode           & 12100, 11700                    \\ 
kids\_allowed           & Yes, No                    \\ 
\hline
\end{tabular}
\caption{Slots appearing in the NLG dataset}
\label{dataset-slots}
\end{table}

\begin{table}[]
\centering
\begin{tabular}{l|l|l|l}
\hline
part                      & Train & Dev & Test \\ \hline
Unique MRs  & 144   & 51  & 53   \\
Corpus size & 3,569 & 781 & 842  \\ \hline
\end{tabular}
\caption{Czech NLG dataset statistics. The unique MRs are counted after delexicalizing the slots.}
\label{dataset-stats}
\end{table}

\subsection{Datasets}

\textbf{Pre-training} We use the Czech-English parallel corpus provided by the WMT 2019 shared task. The dataset comprises of 57 million translation pairs, automatically mined from the web. The data is comprised of a variety of domains (news, subtitles etc). In order to facilitate a fair comparison, we use this corpus for our unsupervised pre-training baselines as well. This effectively results in 114 million monolingual sentences, equally split between English and Czech. \newline
\textbf{NLG} We use the recently released Czech Restaurant dataset, consisting of roughly 3500 examples for training. Further data related statistics can be found in Table \ref{dataset-stats}. The delexicalized MRs in the test set never appear in the training set. As a result, models must learn to generalize to MRs with unseen slot and intent combinations. Table \ref{dataset-slots} lists all the slots that appear in the dataset, along with examples.

\subsection{Training details}
For NMT and MASS, we train transformer models with 93M parameters (6 layers, 8 heads, 512 hidden dimensions). They are trained on a TPU for 1 million steps with Adam optimizer and a learning rate schedule of (1,4K) \footnote{The shorthand form (1.0, 4K) corresponds to a learning rate of 1.0, with 4000 warm-up steps for the schedule, which is decayed with the inverse square root of the number of training steps after warm-up.}. The effective batch size is 1024. \par
For NLG, all our models are trained synchronously on 8 P100 GPUs for 10K steps with a batch size of 32 per GPU. 
We do not perform any hyperparameter tuning. Decoding is performed using beam search, with a beam width of 8. \par
All the transformer based models are implemented in the Lingvo framework \citep{shen2019lingvo} based on Tensorflow \citep{abadi2016tensorflow}. 
fThe best checkpoints are selected based on validation set BLEU score.  

\subsection{Data pre-processing}
Our vocabulary consists of a  sentencepiece model with 32,000 tokens \citep{kudo2018sentencepiece} shared between English and Czech. It is computed on  English and Czech sentences from the pre-training corpus. In order to facilitate a fair comparison, we maintain the same vocabulary across all the transformer based models and baselines. Relying on sentencepieces also ensures that out-of-vocabulary tokens will not be encountered. No special rules or pre-processing is done to tokenize the structured data - we simply feed it as a plain string. 
The input sequence is pre-pended with a task specific token -  \texttt{[TRANSLATE]} for translation, \texttt{[GENERATE]} for NLG. Following \citet{aharoni2019massively}, we pre-pend a second token to specify the desired output language - \texttt{<2en>} for English and \texttt{<2cs>} for Czech \footnote{\texttt{<2en>} is required for the bidirectional NMT model.}.

\begin{table*}[t]
\begin{tabular}{llllllll}
\hline
model                           & BLEU $^\ddagger$ $\uparrow$     & SER $\downarrow$              & NIST $\uparrow$                & METEOR $\uparrow$               & ROUGE-L $\uparrow$              & CIDEr $\uparrow$               & BLEU $^\mathsection$ $\uparrow$ \\ 
\hline
tgen-sota $^\dagger$            & 20.6                            & 2.75                          & 4.77                           & 23.32                           & 42.95                           & 2.18                           & 21.96                              \\
scratch                         & 11.19                           & 63.18                         & 3.06                           & 15.79                           & 28.27                           & 0.84                           & 11.66                              \\
mass                            & 16.61                           & 24.82                         & 4.22                           & 21.16                           & 38.94                           & 1.75                           & 17.72                              \\
nmt                             & 24.41                           & 2.38                          & 5.19 & 25.46                           & 46.85                           & 2.55                           & 25.84   \\
\textbf{binmt}              & \textbf{24.87}                    & \textbf{1.9}                        & \textbf{5.24} & \textbf{25.81}  & \textbf{47.07}    & \textbf{2.60} & \textbf{26.35}  \\ 
\hline
\end{tabular}
\caption{Results. $\uparrow$ implies higher is better, while $\downarrow$ arrow implies lower is better. $\dagger$ We compute BLEU and SER metrics on outputs provided to us by the authors. The other metrics are taken from the paper \citep{duvsek2019neural},
$^\ddagger$ is \textsl{sacrebleu}, $^\mathsection$ bleu as computed by the e2e-metrics suite\footref{e2e-metrics}.}
\label{results-main}
\end{table*}

\subsection{Metrics}
We use BLEU~\citep{papineni2002bleu} as one of our automatic metrics~\footnote{Computed by \textsl{sacrebleu} \citep{post2018call}}. We compute a Slot Error Rate (SER) metric to gauge how well the generated text reflects the structured data. We calculate how many of the slot values in the structured data have been mentioned in the generated text. An example is marked as correct only if all the slot-values in the structured data are present in the output \footnote{SER can be reliably computed only for delexicalizable slots. As a result, the \textsl{kids\_allowed} slot is ignored.}. We refer the reader to the supplementary  material for the exact SER algorithm. We also use the suite of word-overlap-based automatic metrics from the E2E NLG Challenge \footnote{https://github.com/tuetschek/e2e-metrics \label{e2e-metrics}}, supporting NIST \citep{doddington2002automatic}, ROUGE-L \citep{lin2004rouge}, METEOR \citep{lavie2007meteor}, CIDEr \citep{vedantam2015cider} and BLEU. \footnote{Note that this is computed differently from \textsl{sacrebleu}.}

\section{Results and Discussion} 

\subsection{Main Results} \label{results-and-discussion}
We report results in Table \ref{results-main}. Recall that these are models are trained to generate fully lexicalized output. \par
The \textsl{scratch} 
baseline performs quite poorly. While unsupervised transfer learning (\textsl{mass}) performs better, pre-training via machine translation gives the best results by large margin. \textsl{nmt} brings down the SER to just 2.38, a 20 point gain over \textsl{mass}, while improving the BLEU score by 8 points. Similar trends are observed in the other metrics as well. \textsl{binmt} slightly outperforms \textsl{nmt} and leads to further gains across all metrics. These results give credence to our hypothesis that machine translation can be a strong pre-training objective for data-to-text generation in non-English languages. \par
Compared to the pipelined \textsl{tgen-sota} system, both \textsl{nmt} and \textsl{binmt} compare favorably, showing improvements on all metrics, including a 4 point improvement in BLEU. In section \ref{pipelined}, we discuss this result in detail, along with a comparison of the two approaches.


\subsection{Human Evaluation}
We also conduct human evaluations on a set of 200 examples randomly sampled from the test set. Concretely, we measure three metrics - accuracy, fluency and pairwise preference. \par 
\textbf{Accuracy}: Human raters are shown the gold text and the predicted text and are instructed to mark the generated text as inaccurate if any information contradicts the gold text. This effectively catches errors due to hallucinations, incorrect grounding etc. Each example is rated by 3 raters, and we consider an example to be correct if at least two raters say so. \par
\textbf{Fluency}: We show the predicted text to raters and ask them how natural and fluent the text sounds on a 1-5 scale, with 5 being the highest score. Again, each example is rated by 3 raters. We average the scores across all the ratings to get the fluency score. \par
We conduct accuracy and fluency evaluations for our best model (\textsl{binmt}) and the best lexicalized baseline, \textsl{mass}. 
Results are reported in table \ref{results-human}. In terms of fluency, we note that \textsl{mass} produces quite fluent text, likely due to its strong language model. It would seem that unsupervised learning on unlabeled data is enough to generate fluent text, echoing findings of past work \citep{radford2019language}. \textsl{binmt} performs slightly better with a score of 4.83. However, when it comes to accuracy, our model gets a high score of 97.5, surpassing \textsl{mass} by 7.5 points. We take this as proof that transfer learning from machine translation helps produce text that is not only fluent, but much better grounded in the structured data. \par 
\textbf{Pairwise Preference}: We do a side by side evaluation of the predictions from \textsl{binmt} with the gold text written by humans. We show both texts to the raters and ask them which one they prefer on a 7 point Likert scale.
Each example is rated by 3 raters, with the final rating obtained via majority vote. \par 
In 40\% of the cases, our model produces output that is as good as human written text, while in another 30\% there is no majority. Strikingly, in 21\% of the cases, the raters actually preferred the model's output over the human written gold text. The human text is preferred in only 9\% of the cases. These results strongly point to the applicability of this approach to real-world NLG systems.

\begin{table}[]
\centering
\begin{tabular}{l|l}
\hline
  rating             &   percentage     \\ \hline
 much better     & 0.5\%  \\
 better          & 12.5\% \\
 slightly better & 8\%    \\
 about the same  & 40\%  \\ 
 slightly worse  & 2.5\%  \\ 
 worse           & 6.5\%  \\ 
 much worse      & 0\%    \\
 no majority     & 30\%  \\ \hline
\end{tabular}
\caption{Ratings of machine generated output when compared to human written gold text.}
\end{table}


\begin{table}[]
\centering
\begin{tabular}{l|l|l}
\hline
model & accuracy $\uparrow$ & fluency $\uparrow$ \\ \hline
binmt & 97.5     & 4.83    \\ 
mass  & 90       & 4.77    \\ \hline
\end{tabular}
\caption{Human evaluations for accuracy and fluency}
\label{results-human}
\end{table}

\subsection{Low-resource machine translation} \label{low-nmt}
Our previous experiments use NMT models trained on a fairly large corpus. However, for many languages, the amount of available parallel data can be small. Unfortunately, we do not know of any public data-to-text datasets for actual low resource languages. Therefore, to study the impact of the size of bitext corpus, we run experiments in a simulated low-resource setting. We train bidirectional machine translation models on 10\% (5.7 million examples, medium resource, denoted as \textsl{binmt-5m}) and 1\% (570K examples, low resource, denoted as \textsl{binmt-500k}) of the data and use them for fine-tuning the NLG task.  \par
First, to get an idea of how the corpus size effects translation performance, we compute BLEU  scores of each model on the WMT 2019 English-Czech validation set. The medium resource model appears to be as good as the high resource model, but the low resource model is considerably weaker. \par
Next, we fine-tune each of these models on the data-to-text task. From the results in Table \ref{results-nlg}, we see that while the high resource model performs the best, the medium resource models is not far behind in terms of BLEU. Both the high and medium resource models have a comparable SER. Even the low resource model, pre-trained on just 1\% of the translation corpora is significantly better than \textsl{mass}, which has been pre-trained on almost 1.6 billion tokens. The results indicate that machine translation based transfer learning can be successfully applied even when the size of parallel corpus is small, and thus holds promise for low-resource languages.

\begin{table}[]
\centering
\begin{tabular}{l|l}
\hline
Model & BLEU $\uparrow$     \\ \hline
binmt-50m             & 20.95    \\
binmt-5m             & 20.46  \\
binmt-500k             & 15.86 \\ 
\hline

\end{tabular}
\caption{Czech translation performance on the WMT 2019 development set.}
\label{results-nmt}
\end{table}

\begin{table}[]
\centering
\begin{tabular}{l|l|l|l}
\hline
Pre-train & Model & BLEU $\uparrow$  & SER $\downarrow$   \\ \hline
1.6B & binmt-50m             & 24.87 & 1.9   \\
160M & binmt-5m             & 22.17 & 1.43  \\
16M & binmt-500k             & 21.27 & 12.47 \\ 
1.6B & mass                & 16.61 & 24.82 \\
\hline

\end{tabular}
\caption{NLG fine-tuning with low-resource NMT. The first column indicates the number of tokens used for pre-training.}
\label{results-nmt}
\end{table}

\subsection{Low resource NLG} \label{low-nlg}
In this section we study the effects of transfer learning when the size of the fine-tuning corpus is small. We create two random subsets from the NLG training data of size 100 and 1000. Results are reported in Table \ref{results-nlg}. We find that once again, NMT offers substantial gains over MASS.
When fine-tuning on 1000 examples, pre-training with NMT is substantially better (20\% improvement on SER, +3 on BLEU) than fine-tuning MASS with the \textit{full} dataset.
Remarkably, with just 100 examples, our model outperforms  training from scratch on the entire training set by over 3 BLEU, while reducing SER by over 30 points. These results lead us to believe that NMT pre-training can lead to substantial cost savings with respect to training data annotation.

\begin{table}[]
\centering
\begin{tabular}{l|l|ll} 
\hline
Training Size    & Model & BLEU $\uparrow$  & SER $\downarrow$   \\ \hline

& scratch                   & 2.83  & 78.5 \\ 
100 & mass                      & 4.42  & 78.74 \\ 
& binmt                              & 14.62 & 31.82 \\ \hline

& scratch                   & 6.93  & 70.19 \\ 
1000 & mass                      & 9.07  & 66.15 \\ 
& binmt                             & 19.89 & 4.51  \\ \hline

& scratch         & 11.19 & 63.18  \\ 
Full & mass    & 16.61 & 24.82 \\ 
& binmt & 24.87 & 1.9 \\ \hline
\end{tabular}
\caption{Experiments with low-resource NLG}
\label{results-nlg}
\end{table}

\subsection{Out-of-Vocabulary Slot Values}
For real world systems, generalizing to out-of-vocabulary OOV slot-values is essential. However, since NLG datasets are small, this is a major failure mode for models producing lexicalized outputs. The model simply does not see enough unique slot values during training. \par
Generalization to OOV slot-values is hard to measure on this test set, since most of the slot-values already appear in the training set. Therefore, we design a new test set of meaning representations which exclusively contain OOV slot-values. This set contains 100 meaning representations (MRs) with a total of 200 slots and 155 unique OOV slot values. We use Google Search and Wikipedia to sample new values for slots like name, area, food etc. The exact dataset creation procedure is described in the supplementary materials.  \par 
Since we do not have gold text for these MRs, we manually rate the predictions of \textsl{binmt} and \textsl{mass} on this new test and compute slot specific error rates. For each of the 200 slots, we mark it as incorrect if the corresponding slot value is missing from the prediction. \par

\begin{figure}[h]
\noindent\includegraphics[width=0.5\textwidth]{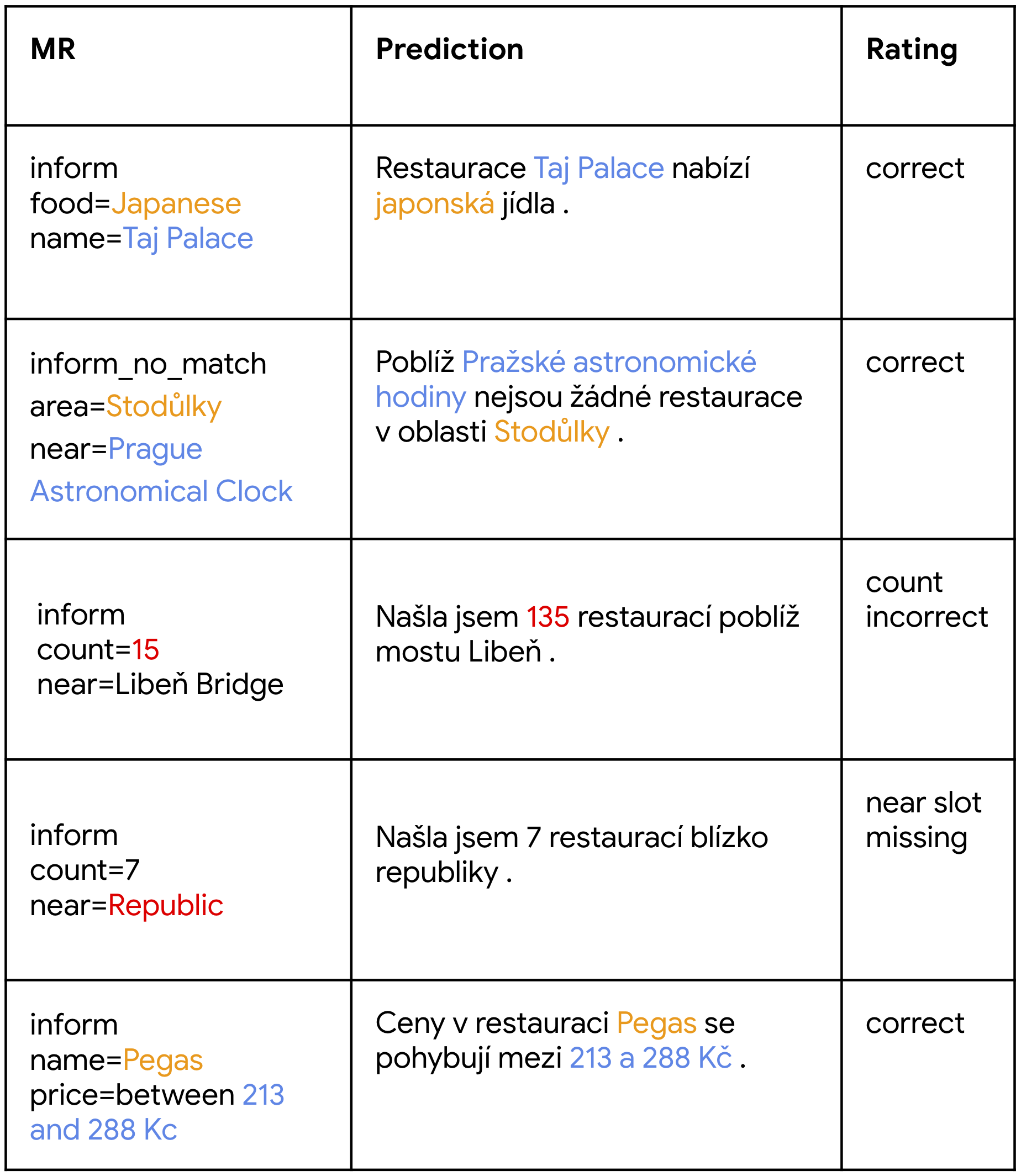}
\caption{
Sample predictions of \textsl{binmt} on the OOV test set. Aligned segments from the structured data and natural language have the same color.
}
\label{fig:oov}
\end{figure}

\begin{table}
\centering
\begin{tabular}{l|l|l|l|l}
\hline
slot    & unique & total & errors &  accuracy \\ 
\hline
 name                & 39                     & 70             & 3  & 95.7                          \\
area                           & 30                     & 30             & 2 & 93.3                            \\
address                          & 10                     & 10             & 1     & 90.0                        \\
food                            & 10                     & 10             & 4            & 60.0                \\
phone                 & 10                     & 10             & 0         & 100.0                   \\
count                          & 20                     & 20             & 6       & 70.0                      \\
post code                       & 10                     & 10             & 0            &100.0                \\
price                            & 10                     & 10             & 0  & 100.0                          \\
near                            & 16                     & 30             & 1          & 96.7                  \\
\hline
total          & 155           & 200   & 17   & 91.5 \\
\hline
\end{tabular}
\caption{Out-of-Vocabulary test set. \textsl{unique} refers to the number of unique values the slot takes in the test set. \textsl{total} is the number of times the slot appears.}
\label{results-oov}
\end{table}

Unsupervised pre-training through MASS completely fails to generalize to OOV values - none of the slots are realized in the predictions. Looking at the output, we noticed that \textsl{mass} has a strong tendency to hallucinate or simply output slot values seen during training. The poor performance further reinforces the practical popularity of delexicalized models and highlights the need for challenging OOV test sets. \textsl{binmt} on the other hand, remains robust - 91.5\% of the slots from the MR are realized in the predictions. We show some examples in Figure \ref{fig:oov}, along with slot specific scores in Table \ref{results-oov}. The results confirm that NMT can greatly improve the robustness to unseen values. Further reducing the performance gap between seen and unseen values is an important area of future work.


\subsection{Comparison with pipelined approaches.} \label{pipelined}
We showed in section \ref{results-and-discussion} that our lexicalized NMT based models compare favorably to the current best model for this dataset - the \textsl{tgen-sota} system \citep{duvsek2019neural}. Note that our aim in this work is not to beat state-of-the-art, but to gauge the effectiveness of machine translation as a pre-training strategy for NLG. Nevertheless, a comparison with  \textsl{tgen-sota} offers some interesting insights. We first describe \textsl{tgen-sota} in detail. It consists of the following pipelined components: \par
\textbf{Delexicalized Generator}: A TGen model trained to generate delexicalized output, either as a sequence of words, or as a sequence of interleaved lemmas and morphological tags. \par
\textbf{Classifier Reranker}: An LSTM based NLU classifier that ranks top-k beam search hypotheses. The classifier is trained to predict the MR from the text and can be used to select outputs that are most faithful to the meaning representation. \par
\textbf{LM Lexicalizer}: Due to heavy inflections, inserting slot-values into the placeholders verbatim leads to ungrammatical texts. To remedy this, the authors train a language model, which picks the most probable surface form for every slot-value. \par
The delexicalized generator and reranker ensure that the generated text is firmly grounded in the MR. 
And while the language model does a good job of selecting the correct surface form, the technique relies on the existence of an exhaustive list of surface forms associated with each slot-value. \par 
Such a list can be hard to obtain and maintain. The problem is exacerbated for open-domain slots like movies, people, restaurant names which can take a large number of values and are constantly expanding. In addition, since several slot values in the structured data are in English, some form of bilingual knowledge (eg - dictionaries) is also necessary. The linguistic expertise and resources required to create such a database, or even to train alternative models that inflect text based on morphological tags, may not be easily available (especially for low resource languages). Finally, pipelined methods require training, tuning and maintaining separate models for each component. \par
In stark contrast, our approach is completely end-to-end, consisting of a single model which directly produces fully lexicalized outputs without relying on any linguistic resources. The only dependence is on availability of parallel data and as we show in section \ref{low-nmt}, we can learn accurate models even in low resource NMT settings. Recent advances in bitext mining have resulted in sizeable bitext corpora for many low resource language pairs \citep{schwenk2019ccmatrix, schwenk2019wikimatrix}, bolstering the potential use of this approach. Finally, as we show in the next section, NMT pre-training can also be used to develop improved delexicalized models and subsequently be incorporated into the pipelined approach.

\subsection{Delexicalized NLG} \label{delex-nlg}
In this set up, the model must produce delexicalized output. This is achieved by replacing certain slots in the output by placeholders, similar to the example in Figure \ref{fig:delex}. The model produces output with these these placeholders, which are subsequently filled in as a separate step. The advantages of training delexicalized models include robustness to out of vocabulary values for slots involving  entities. Such two step methods are common in practice. Producing delexicalized text is arguably a simpler problem, since the model needs to just output placeholders instead of fully lexicalized text. Every slot except the binary slot \textsl{kids\_allowed} are delexicalized. \par
We compare NMT with with MASS and a strong TGen based delexicalized model proposed by \citet{duvsek2019neural}. From the results in Table \ref{results-delex}, we see that all the models exhibit low SER, as expected. Our model outperforms the best baseline by 5 BLEU points, pointing to the applicability of our approach even in the case of delexicalized NLG. We leave the combination of this model with a lexicalizer component for future work.

\begin{table}[]
\centering
\begin{tabular}{l|ll} 
\hline
model    & BLEU $\uparrow$  & SER $\downarrow$   \\ \hline
baseline-mass    & 23.48 & 1.07  \\ 
binmt            & 30.87 & 0.95  \\ 
tgen-delex$\dagger$   & 25.34 & 1.22  \\ \hline
\end{tabular}
\caption{Results on delexicalized NLG. \newline
$\dagger$ We compute \textsl{sacrebleu} on outputs provided to us by the authors \citep{duvsek2019neural}.}
\label{results-delex}
\end{table}

\section{Conclusion and Future Work}
In this work we investigated neural machine translation based transfer learning for data-to-text generation in non-English languages. Using Czech as a target language, we showed that such an approach is effective and surpasses the performance of unsupervised transfer learning. It enables us to learn simple, fully lexicalized end-to-end models that perform on par with a sophisticated, linguistically informed pipelined system. Experimental results suggest several desirable properties including improved sample efficiency,  robustness to unseen values and potential applications to low resource languages. At the same time, the approach can also be leveraged to improve performance of delexicalized models. \par
Studying pre-training on a wide variety of languages, especially those with different scripts, is a direct line of future work. Since this is mainly hindered by a lack of datasets, we hope to develop data-to-text corpora for other languages, including ones that are truly low-resource. 


\section*{Acknowledgments}
We would like to thank Markus Freitag for insightful discussions and Ondřej Dušek for providing the \textsl{tgen-sota} model outputs.

\bibliography{anthology,acl2020}
\bibliographystyle{acl_natbib}

\end{document}


\maketitle

\section{Supplementary Material}

\subsection{OOV Dataset Creation}\label{oov-dataset}
Here we describe the process of creating a challenge set to test out-of-vocabulary generalization.
\begin{itemize}
    \item Sample 10 meaning representations (MR) containing open value slots.
    \item Construct a list of slot values for each open value slot via Google Search and Wikipedia. Slots like \textsl{area}, \textsl{address} etc. are localized i.e. they are from Czech Republic. Numeric slots like \textsl{count} and \textsl{phone number} are automatically synthesized.
    \item For each MR, randomly sample slot values from the above list, resulting in a MR that contains only OOV slot values. Synthesize 10 such examples for each MR. This results in a total of 100  MRs.
\end{itemize}

\subsection{Computing Slot Error Rate (SER)} \label{computing-ser}
Conceptually this can be achieved by string matching between the slot values and generated text, but such string matching does not work well in the lexicalized setting for non-English languages, since certain slot-values are translated (dinner, breakfast etc), while others are inflected (restaurant names). Fortunately, the authors also provide an exhaustive list of surface forms containing all possible translations and inflections for each slot value. This lets us easily compute the SER \footnote{SER can be reliably computed only for delexicalizable slots. As a result, the \textsl{kids\_allowed} slot is ignored}.